\newif\ifdraft
\draftfalse

\documentclass[twocolumn]{article}
\usepackage{cite}
\usepackage{amsmath,amssymb,amsfonts}
\usepackage{bbm}
\usepackage{graphicx}
\usepackage{hyperref}
\usepackage{caption,subcaption}
\usepackage{multirow}
\usepackage{cleveref}
\usepackage{comment}
\usepackage{hhline}
\usepackage{todonotes}
\usepackage{graphicx}
\usepackage{caption,subcaption}
\usepackage{siunitx}
\usepackage{booktabs}
\usepackage{csvsimple}
\usepackage{datatool}
\usepackage{array}

\usepackage{tikz}
\usetikzlibrary{matrix}
\usepackage{arydshln}

\usepackage[normalem]{ulem}
\newcommand{\stkout}[1]{\ifmmode\text{\sout{\ensuremath{#1}}}\else\sout{#1}\fi}

\ifdraft
\newcommand\removenote[1]{{\color{gray}#1}}
\newcommand\removed[1]{\textcolor{red}{\stkout{#1}}}

\else
\newcommand\removenote[1]{}
\newcommand\removed[1]{}

\fi

\newif\ifSuppressMemo
\ifSuppressMemo
\newcommand{\memo}[1]{}
\else
\usepackage{color}
\newcommand{\memo}[1]{{\bf \textcolor{red}{[#1]}}}

\fi

\newcommand\blfootnote[1]{%
  \begingroup
  \renewcommand\thefootnote{}\footnote{#1}%
  \addtocounter{footnote}{-1}%
  \endgroup
}

\begin{document}

\title{On the Robustness of Pretraining and Self-Supervision for a Deep Learning-based Analysis of Diabetic Retinopathy }
\author{Vignesh Srinivasan, Nils Strodthoff, Jackie Ma,\\ Alexander Binder, Klaus-Robert M\"uller, Wojciech Samek 
\blfootnote{VS, NS, JM and WS are with the Department of Artificial Intelligence, Fraunhofer Heinrich Hertz Institute, 10587 Berlin, Germany (email: firstname.lastname@hhi.fraunhofer.de).
NS and WS are also with BIFOLD – Berlin Institute for the Foundations of Learning and Data, 10587 Berlin, Germany.
AB is with the Department of Informatics, Oslo University, 0373 Oslo, Norway. 
KRM is with the Machine Learning Group, Technische Universit{\"a}t Berlin, 10587 Berlin, Germany, and also with the Department of Artificial Intelligence, Korea University, Seoul 136-713, South Korea, the Max Planck Institute for Informatics, 66123 Saarbr{\"u}cken, Germany, and BIFOLD – Berlin Institute for the Foundations of Learning and Data, 10587 Berlin, Germany. (e-mail: klaus-robert.mueller@tu-berlin.de).
}}
\date{}
\maketitle


\maketitle

\begin{abstract}
There is an increasing number of medical use-cases where classification algorithms based on deep neural networks reach performance levels that are competitive with human medical experts. To alleviate the challenges of small dataset sizes, these systems often rely on pretraining. In this work, we aim to assess the broader implications of these approaches. 
For diabetic retinopathy grading as exemplary use case, we compare the impact of different training procedures including recently established self-supervised pretraining methods based on contrastive learning. To this end, we investigate different aspects such as quantitative performance, statistics of the learned feature representations, interpretability and robustness to image distortions. 
Our results indicate that models initialized from
ImageNet pretraining report a significant increase in performance, generalization and robustness to image distortions. In particular, self-supervised models show further benefits to supervised models. 
Self-supervised models with initialization from ImageNet pretraining not only report higher performance, they also reduce overfitting to large lesions along with improvements in taking into account minute lesions indicative of the progression of the disease.  
Understanding the effects of pretraining in a broader sense that goes beyond simple performance comparisons is of crucial importance for the broader medical imaging community beyond the use-case considered in this work.

\end{abstract}

\section{Introduction}
The role of computer vision algorithms based on deep learning in medical imaging in the form of decision support systems has increased steadily in the past few years \cite{raghu2019transfusion, gulshan2016development, voets2019reproduction, sowrirajan2020moco, ChenKSNH20, azizi2021big, binder2021morphological}. There is an enormous amount of data that is being produced on a daily basis from different areas using different imaging modalities such as MRI, CT, microscopy, etc., leading to an unprecedented potential for machine learning algorithms. However, while there exists a lot of data, it is usually not prepared to be used for research in machine learning. In particular, it is often unlabeled as the labeling process is expensive and time-consuming, or sometimes medical experts may not agree on the appropriate label.

A practitioner
using Deep Neural Networks (DNN)
for the task of medical imaging,
is faced with
a plethora of options when it comes to the training methodology for the DNN. 
Several factors can influence the decision making process
including, but not limited to 
the size, noise level and quality of the 
dataset at hand, 
computational resources
available and 
robustness of 
the trained DNN.
Transfer learning 
for medical imaging
from models trained
on natural images
have been found to be
beneficial for improvements in performance
along with 
speeding up convergence
\cite{raghu2019transfusion, NeyshaburSZ20}. 
A straightforward way of 
utilizing transfer learning 
is to finetune a model that is initially trained 
on ImageNet \cite{imagenet_cvpr09}
on  the  medical  dataset.

\begin{figure*}[ht]
         \centering
         \includegraphics[width=\textwidth,keepaspectratio]{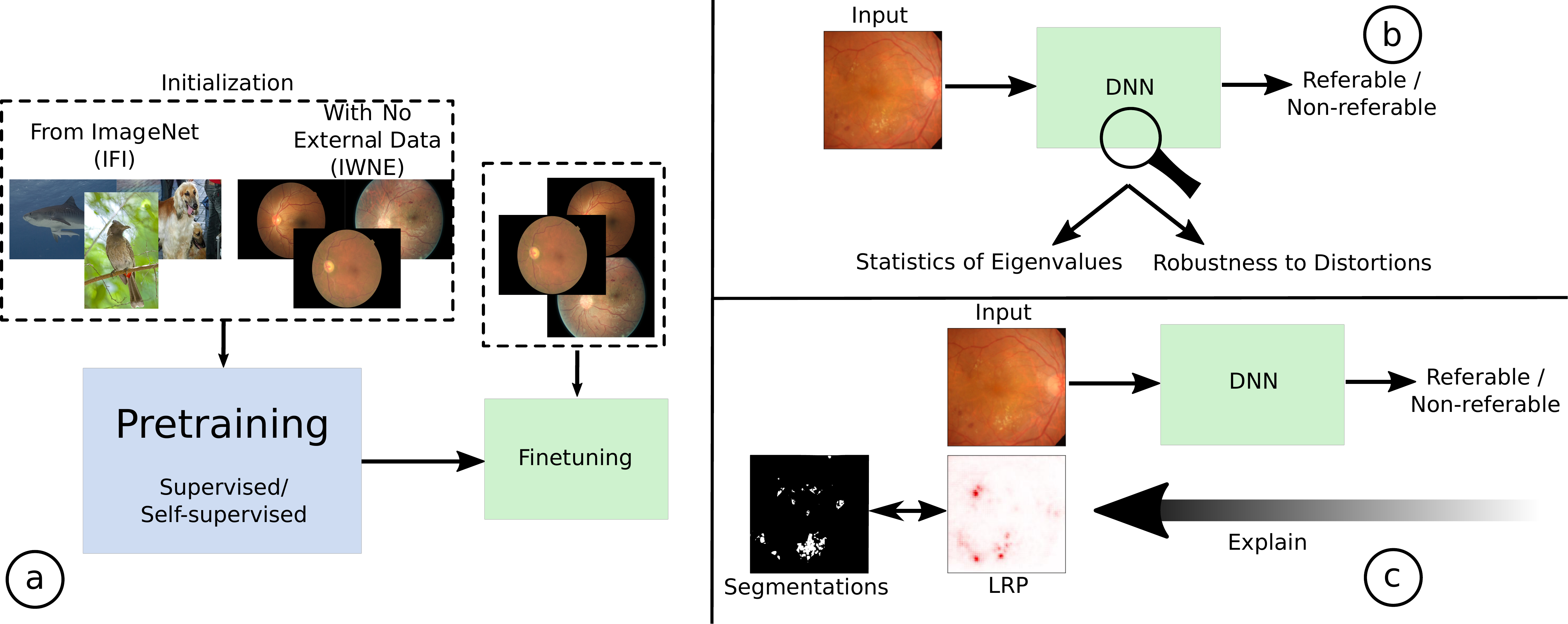}
         \caption{Overview of the experiments presented in this work. 
         a) shows the different training strategies which includes pretraining and finetuning. 
         b) investigates the statistics of the eigenvalues of the feature representations learned by the different methods which lead to increased robustness to distortions.
         c) shows the experiments we perform using the Indian Diabetic Retinopathy Image Dataset (IDRiD) challenge data \cite{porwal2020idrid} to quantitatively evaluate the cues learned. 
         }
\end{figure*}

Other common state-of-the-art methods in machine learning are \textit{supervised-learning} methods, i.e.\ models that are trained with labeled data, opposed to other methods that require only some or even no labeled data such as \textit{semi-supervised} or \textit{self-supervised learning}. Fortunately, the field of self-supervised learning has recently advanced significantly \cite{he2020momentum, chen2020simple, grill2020bootstrap,caron2021unsupervised}, which gives rise to hope for a successful deployment of machine learning in medical applications without relying on overly large amounts of labeled data. A first result in this regard was obtained in \cite{anonymous2021mocopretraining, azizi2021big, sriram2021covid} where the authors showed that  pretraining using self-supervision helps to improve the models for chest x-ray classification \cite{irvin2019chexpert}, dermatology condition classification \cite{liu2020deep} and Covid-19 deterioration prediction \cite{sriram2021covid}.

With widespread adoption of 
transfer learning in medical imaging, 
it becomes essential to explore 
the differentiating features
of the various training methodologies---supervised or self-supervised. 
While \cite{raghu2019transfusion}
observe the effects of pretraining 
in supervised learning on
the speed of convergence 
and feature representations,
\cite{NeyshaburSZ20} 
study the effects on the performance
of pretrained models from ImageNet
providing improvements on 
diverse datasets and the
quality of the features learned.
Despite the benefits of 
transfer learning, 
it has, however, remained unclear what 
transfer learning, especially 
with self-supervised learning 
actually exploits when making a prediction. 
For this (as we will see) simply 
looking at  performance metrics like 
classification accuracy or area under the 
operating curve (AUC) is not sufficient.
The potential advantages of using self-supervised 
methods over supervised methods 
for medical imaging 
beyond such performance metrics
thus remain a challenging object of study.

In this contribution, we demonstrate for diabetic retinopathy (DR) as a particular medical imaging use case, that going beyond metrics of predictive performance is mandatory. We further analyze robustness to statistical variations of the data. Furthermore we validate previous results on smaller data sets which are of ubiquitous interest to practitioners in medical data science.

To this end, 
we perform a detailed 
study of what is being 
learned by the 
different training 
methodologies available
to train a DNN for 
medical imaging. 
Broadly, the training methodologies will 
be categorized into two types: 
\begin{itemize}
    \item Fully supervised (FS)
    \item Self-supervised with contrastive learning (CL)
\end{itemize}
along with two types of initialization of the weights before training on the medical dataset:
\begin{itemize}
    \item Initialization with no external data (IWNE)
    \item Initialization from ImageNet (IFI)
\end{itemize}
The focus of this paper is to
study the effects of 
training the DNN using
these strategies
and evaluate the benefits.
Our contributions are as follows: \\
\textbf{1)} We evaluate the performance of the
four different training strategies:
supervised and self-supervised models 
using models trained with or without using external data for 
pretraining 
for detecting diabetic retinopathy 
in retinal images. 
We find that IFI helps in achieving 
significant gain in performance, especially when a limited 
amount of the downstream (medical) labeled dataset is used.
IFI-CL provides a further increase in performance.
\\
\textbf{2)} Given that IFI is beneficial
in terms of performance, 
we investigate 
what makes them better by analyzing the 
eigenvalue spread of the 
activations on the hidden layers. 
We find that the 
redefined conditioning number 
for the IFI models is lower than 
that of IWNE models
for the initial layers
that are important for learning 
diverse and effective feature
representations from the input. 
IFI makes
the eigenvalue spread of the 
activations of the first hidden layer broader
implying wider range kernels firing
for a given input. 
In both IWNE 
as well as IFI models, 
we show that CL 
achieves broader eigenvalue spread
compared to its supervised counterparts.
\\
\textbf{3)}
Using explainability of DNNs, 
we investigate what the different models 
look at in the input for making a decision. 
With the help of ground-truth 
segmentation maps available for 
diabetic retinopathy
on the IDRiD challenge \cite{porwal2020idrid},
we 
study  
 in a quantitative manner what
 information was used by the models to make the prediction.
We find that IWNE-FS overfits to
large lesions like hard exudates and ignores smaller lesions
to predict the disease.
IFI models show significantly reduced 
tendency to overfit to one particular type of lesions.
Especially IFI-CL is able to consider 
a wider range of lesions to make an accurate prediction
for the disease.
\\



\begin{figure*}
    \centering
    \begin{subfigure}[t]{0.49\textwidth}
        \centering
        \includegraphics[width=.8\textwidth]{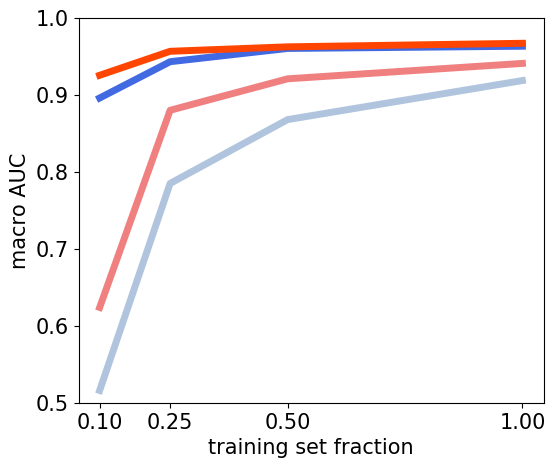}
        \caption{Training and evaluation on Eyepacs-1}
    \end{subfigure}%
    ~ 
    \begin{subfigure}[t]{0.49\textwidth}
        \centering
        \includegraphics[width=0.8\textwidth]{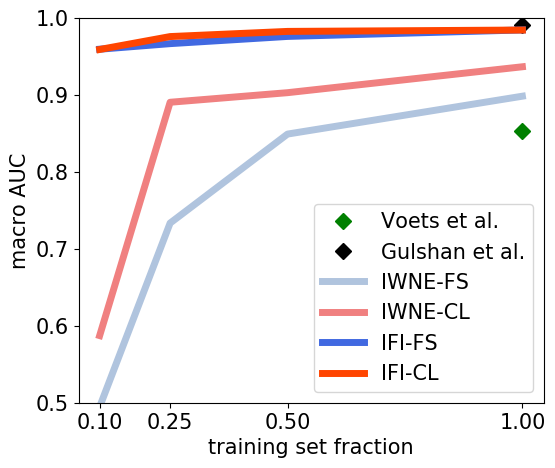}
        \caption{Training on Eyepacs-1 and evaluation on Messidor-2}
    \end{subfigure}
   
    \caption{Classification performance on Eyepacs-1 and Messidor-2 dataset for referrable DR as a function of fraction of the downstream training set used for training for four different training procedures. 
    The state-of-the-art method for DR---Voets et al. \cite{voets2019reproduction} and Gulshan et al. \cite{gulshan2016development} are shown as green and black diamonds for training with the full dataset for the Messidor-2 dataset.}
    \label{fig:auc}
\end{figure*}

\section{Related Work}

\textbf{Diabetic Retinopathy:}
DNN has seen wide adoption for the task of DR
in \cite{gulshan2016development, takahashi2017applying, gargeya2017automated, lam2018retinal, lam2018automated, gao2018diagnosis, zeng2019automated, wang2018diabetic, wan2018deep, chen2018detection, johari2018early, xu2018improved, zhang2019automated, grzybowski2020artificial, poplin2018prediction, varadarajan2018deep, bajwa2019combining, rakhlin2018diabetic, voets2019reproduction, ludwig2020automatic} among others.
While some methods train their model 
from scratch \cite{gargeya2017automated, rakhlin2018diabetic, bajwa2019combining, takahashi2017applying,Leibig2017},  
IFI models have predominantly
achieved higher performance 
\cite{gulshan2016development, wan2018deep, voets2019reproduction, lam2018automated, poplin2018prediction, ludwig2020automatic}.
Some methods also perform their training on large private data \cite{gulshan2016development, xu2018improved, zhang2019automated, gao2018diagnosis, takahashi2017applying}. A reproduction study of \cite{gulshan2016development} was performed by \cite{voets2019reproduction} showing difficulty in achieving 
similar performance for DR when trained on publicly available datasets. 
Systematic study of using uncertainty measures for DR were also conducted by \cite{Leibig2017, filos2019systematic}.
While \cite{lam2018retinal}
studied the probability maps 
with ground-truth segmentation maps
to ascertain what the DNN prediction 
was looking for, 
\cite{SAYRES2019552} studied 
a computer-assisted setting with explanation 
methods for deep learning models in 
grading for DR.
There is, however, no dedicated study on the implications of different 
training methodologies.

\begin{table}[!t]
\centering
\begin{tabular}{@{}lll@{}}
\toprule
     Dataset & \# instances & \# patients \\ 
\midrule
EyePacs-1 (EyP) \cite{Eyepacs1} & 88,702 & 44,351 \\
\hline
   Messidor-2 \cite{Messidor2} & 1,744 & 872 \\
\hline
   IDRiD \cite{porwal2020idrid} & 80 &  - \\
\bottomrule
\end{tabular}
\caption{Diabetic retinopathy datasets used for this study.}
\label{tab:datasets}
\end{table}

\textbf{Supervised vs.\ Self-supervised Learning:}
Self-supervised learning has been utilized 
in a wide range of biomedical applications
including 
chest x-rays \cite{sowrirajan2020moco, ChenKSNH20, azizi2021big, sriram2021covid}, 
diabetic retinopathy \cite{TalebLDSGBL20, holmberg2020self}, 
covid detection \cite{sriram2021covid} etc.
In spite of the improvements shown by self-supervised learning, \cite{Geirhos2020similarity} find that self-supervised models behave quite similarly to their supervised counterparts in many aspects of robustness. 
Self-supervised
models report a slightly higher 
performance gain 
over their supervised counterparts 
on medical imaging \cite{sowrirajan2020moco, azizi2021big}.
Recent works show the generalizing capabilities 
of self-supervised learning on chest x-rays \cite{navarro2021evaluating}. 
The improvements and benefits 
still need to be rigorously investigated
to ascertain the limits of 
using self-supervised learning on 
real-life healthcare applications.

\textbf{IWNE vs IFI}
Pretraining on ImageNet dataset (i.e. IFI), either supervised or self-supervised, is considered an effective strategy \cite{hendrycks2019using, hendrycks2019self, HendrycksLWDKS20, Djolonga2020transfer, NeyshaburSZ20, JiangCCW20, chen2020adversarial, ChenKSNH20, azizi2021big, sowrirajan2020moco}.
Several benefits have been attributed to 
pretraining including
robustness \cite{hendrycks2019using, hendrycks2019self, HendrycksLWDKS20, Djolonga2020transfer, NeyshaburSZ20},
to generalization \cite{peng2018using, chen2019med3d} to finding sparser subnetworks from the original \cite{Chen2020Lottery}
and 
to speed up convergence on the downstream task \cite{raghu2019transfusion, NeyshaburSZ20}. 
Using IFI for DR has been widely adopted owing to benefits in performance
\cite{raghu2019transfusion, gulshan2016development, wan2018deep, lam2018automated, rakhlin2018diabetic, voets2019reproduction, kandel2020transfer}.
The performance benefits 
of pretraining have been observed 
even on diverse datasets which 
seem distant from the ImageNet dataset
\cite{NeyshaburSZ20}.
The benefits of pretraining
can be attributed
to effective feature extracting capability 
of pretrained models in the lower layers
\cite{raghu2019transfusion, NeyshaburSZ20}.
Although, it is unclear how this translates to a DNN being 
used for a downstream task.
While the above mentioned methods
investigate supervised learning, 
we make a comparative study of IWNE vs IFI
along with FS vs CL and their combinations
to understand their 
differentiating features.

\section{Materials \& Methods}
\subsection{Datasets}
We focus on diabetic retinopathy (DR) as a use case for our investigations and solely work on publicly available datasets,
which are summarized in Table~\ref{tab:datasets}.

\begin{figure*}
    \centering
    \begin{subfigure}[t]{0.6\textwidth}
    \centering
    \includegraphics[height=2in,keepaspectratio]{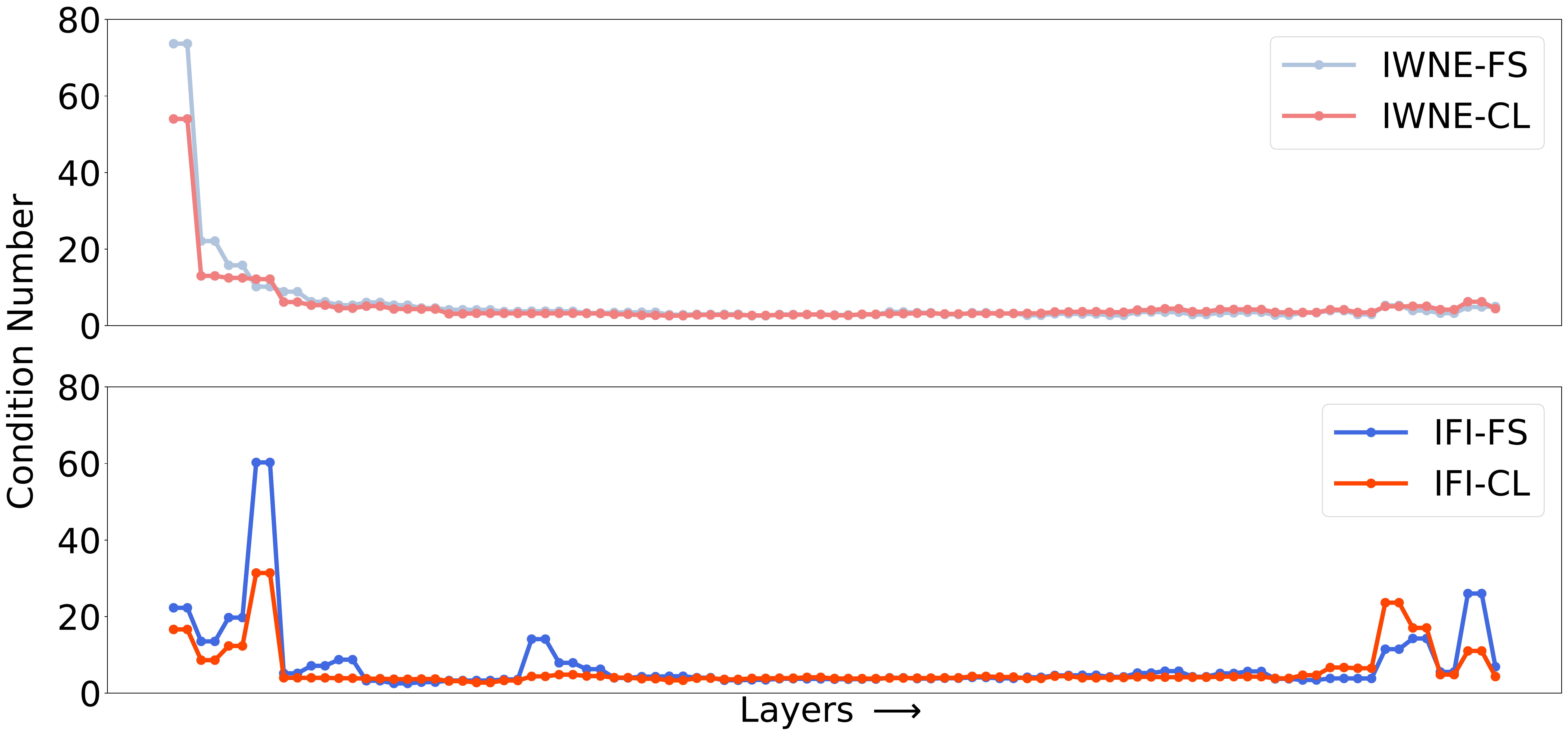}
    \caption{Condition number for each layer of ResNet50 and for the different models.
    }
    \label{fig:condition_number}
    \end{subfigure}
    ~
    \begin{subfigure}[t]{0.37\textwidth}
    \centering
    \includegraphics[height=2in,keepaspectratio]{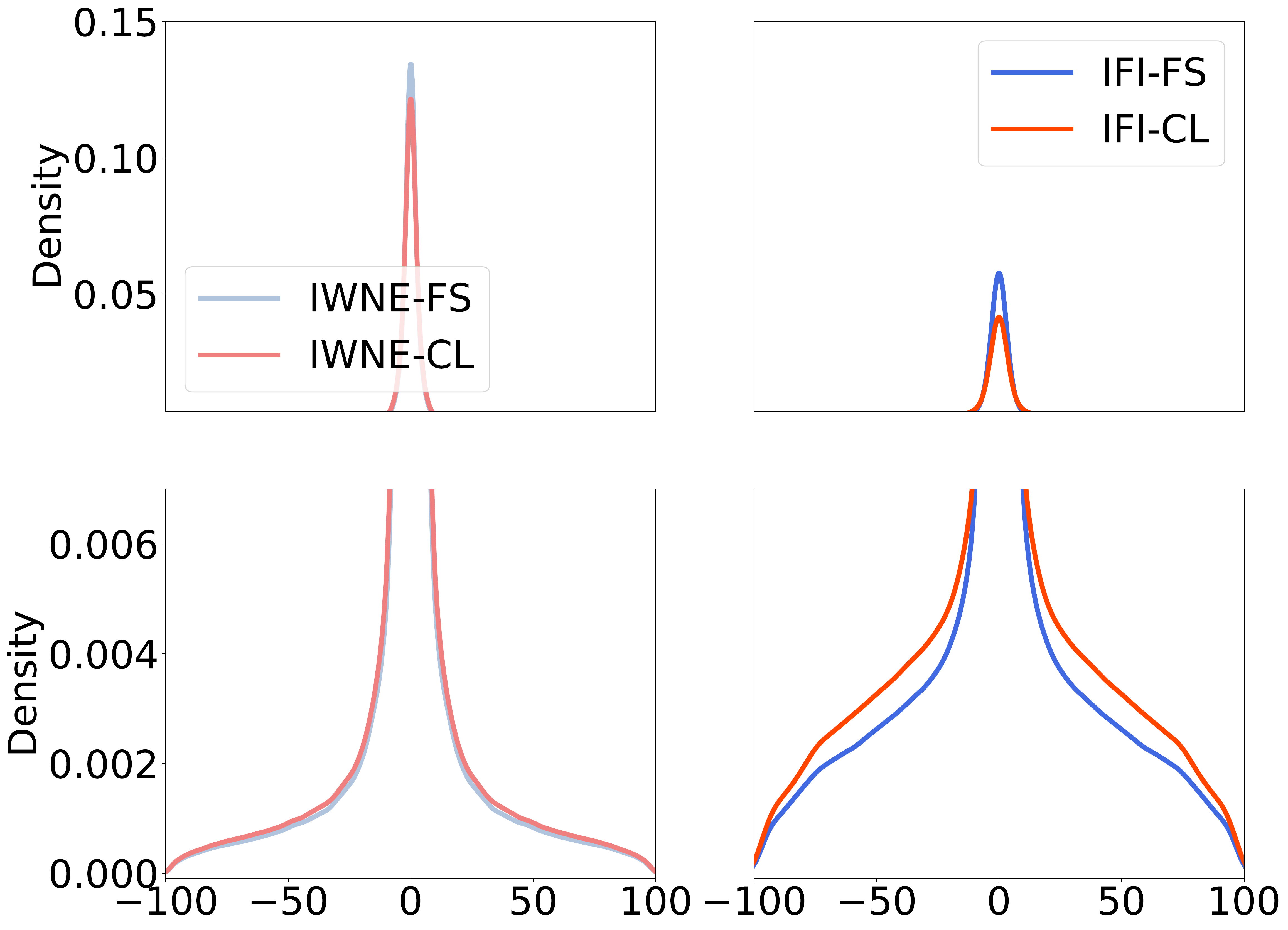}
    \caption{Eigenvalues of the activations of the first convolutional layer 
    made symmetrical around $0$ and plotted in the form of density for better visualization.
    } 
    \label{fig:eigenvalues_conv1}
    \end{subfigure}
    \caption{The statistics of the eigenvalues are shown here.
    a) shows the condition number of all the layers and 
    b) shows the eigenvalues of the 
    activations of the first
    convolutional layer. 
    }
\end{figure*}

We make use of the Eyepacs-1 dataset \cite{Eyepacs1}, which is available from a former Kaggle challenge. 
The images are graded from a scale of $0$ to $4$ (0: no DR, 1: mild DR, 2: moderate DR, 3: severe DR, 4: proliferative DR) according to the
International Clinical Diabetic Retinopathy (ICDR) severity scale.
DR advances from a healthy eye to a proliferate one slowly and may also take years. 
However, this transition is discrete and often goes undetected to worsen into a proliferate DR.
Hence, it is essential that this progression is detected and a timely medical diagnosis is performed.
In our experiments, we train the models to perform the quinary classification using all the five grades. 
During inference, we modify it to a binary classification problem by considering
classes $[0-2]$ as \emph{healthy} and classes $[3-4]$ as \emph{disease}. 
This binary class formulation is consistent with referable DR (rDR) classification in
\cite{gulshan2016development, voets2019reproduction}.


The Eyepacs-1 dataset \cite{Eyepacs1} 
consists of $35216$ images in 
the training set and 
$53576$ in the test set. 
We utilize non overlapping 
set of around $15\%$ of the  
training set 
as the validation set.
We train all our different methods on the training set of Eyepacs-1 dataset
and evaluate the performance of the 
models on two datasets---test set of Eyepacs-1 and Messidor-2 \cite{Messidor2}.
Messidor-2 dataset \cite{Messidor2} is a benchmark dataset consisting of $1744$ images that are $100\%$ gradable. Since the dataset is not used for training and was collected under different conditions at a different geographical location and with different hardware,
the evaluation on the Messidor-2 dataset is supposed to measure the generalization performance of the algorithms. Hence, we use all the images of this dataset for testing. 
We report the AUC for the binary rDR classification task on the respective test sets of each dataset.


\begin{table}
\centering
\begin{tabular}{@{}lll@{}}
\toprule
    Method &  Distribution & Parameters\\
    \midrule
     IWNE-FS & Pareto &  $\alpha=1.45$\\
     \textbf{IWNE-CL} & \textbf{Pareto} &  $\boldsymbol{\alpha}=\boldsymbol{1.28}$\\
    \hline
    IFI-FS & Pareto & $\alpha=0.87$\\
     \textbf{IFI-CL}  & \textbf{Pareto} & $\boldsymbol{\alpha}=\boldsymbol{0.73}$ \\
    \bottomrule
\end{tabular}
\caption{Distribution fitting for the eigenvalues of the activations of the first layer. 
For all the four models, 
the eigenvalues are best 
parametrized by a 
Pareto distribution.
We also find that the
self-supervised models show 
smaller value for the shape
parameter of the Pareto distribution.
}
\label{tab:distfit}
\end{table}

\subsection{Models \& Training Procedures}
\label{sec:models_training}
We compare the four training setups which are eventually trained on the DR target dataset.
\begin{itemize}
    \item \emph{Initialization With No External Data (IWNE)} 
        \begin{itemize}
        \item \textbf{FS}: supervised training on the DR dataset starting from randomly initialized weights. 
        \item \textbf{CL}: self-supervised pretraining on the target domain and finetuning also on the same dataset using labeled data.
        \end{itemize}
    \item \emph{Initialization From ImageNet Data (IFI)}
        \begin{itemize}
        \item \textbf{FS}: supervised training on the DR dataset starting from supervised ImageNet-pretrained weights.
        \item \textbf{CL}: self-supervised pretraining on ImageNet dataset and finetuning on the DR dataset using labeled data.
        \end{itemize}
\end{itemize}
For comparability, we fix the architecture and use a Resnet50 \cite{he2016deep} model for all of our experiments. In the self-supervised setting, 
we pretrain the models using 
MoCoV2 strategy \cite{chen2020improved}.
For the supervised pretraining, 
we use the ImageNet-pretrained model provided by torchvision.
The IWNE models are trained for $500$ epochs with a learning rate of $10^{-4}$. 
Pretrained models have shown to be faster at convergence than the models trained from scratch \cite{raghu2019transfusion, NeyshaburSZ20}. 
Hence, we finetune the IFI models starting from 
ImageNet-pretrained weights for $50$ epochs
with a learning rate of $10^{-3}$. 
The AdamW optimizer \cite{loshchilov2018decoupled} 
with weight decay 
was used in all the settings. 
The best models in each training run was chosen based on the maximum AUC score achieved on the validation set and this model was used for inference on the test.

\section{Experiments \& Results}
\label{sec:experiments}

\begin{figure*}[ht]
    \centering       \includegraphics[height=2.7in,keepaspectratio]{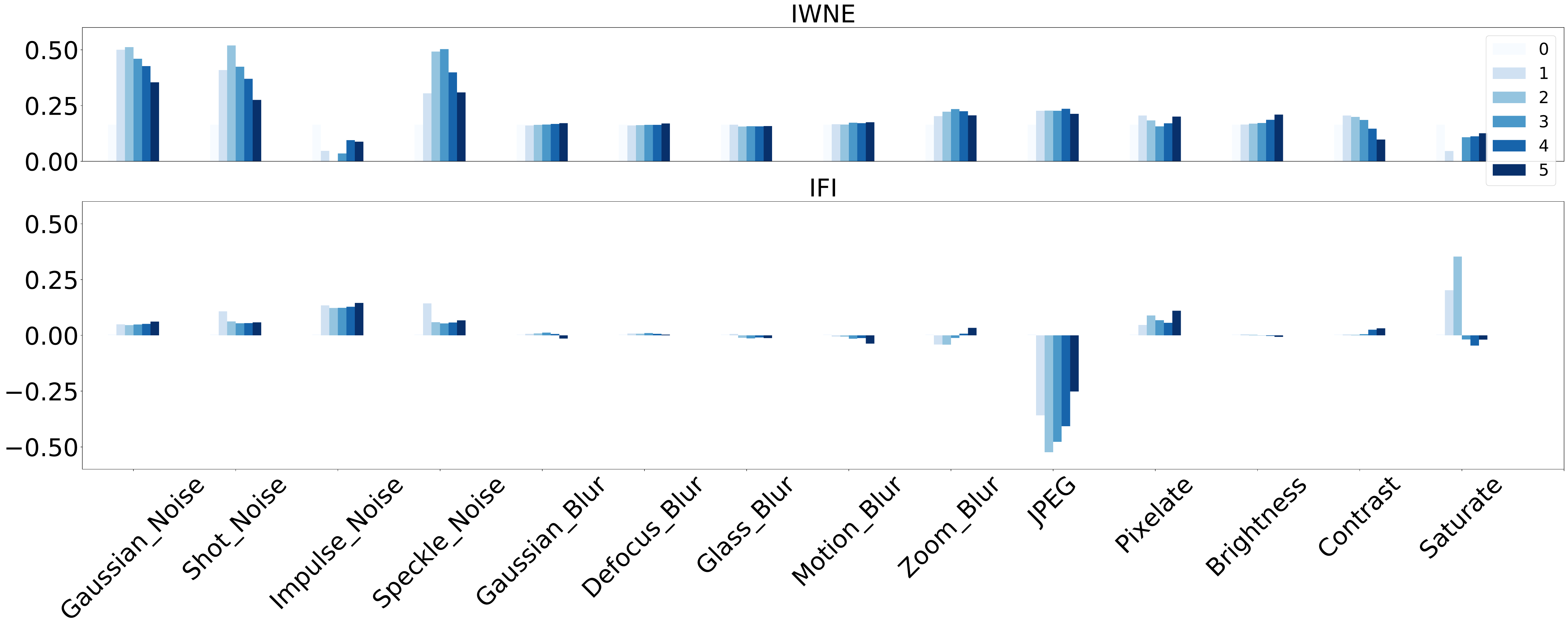}
      \caption{This figure shows the  robustness to distortions for the different models. The difference in the softmax probabilities of the output between the CL and FS model is plotted here. 
      The intensity of the color indicates the severity of 
      the distortions.
      Top row shows the difference for IWNE models. 
      Bottom row shows the difference for IFI models. In case of IWNE, the difference is consistently positive, implying that the self-supervised model has a higher prediction score than the plainly supervised model and thus exhibits a higher robustness to distortions.
      See Section \ref{ssec:robustnesstodistortions} for a detailed discussion.
      }
      \label{fig:robustness}
\end{figure*}

\subsection{Quantitative performance}
We evaluate the performance of the different methods discussed in Section~\ref{sec:models_training}
in terms of AUC. 
Each model was trained on the full dataset and on various fractions of the training set down to a fraction of $10\%$ labeled samples. 
Figure~\ref{fig:auc} shows the final AUC of the binary classification for rDR. We find largely consistent results in terms of the ranking and overall behavior of the different training procedures between evaluation on a subset of the Eyepacs-1 dataset used for training and an evaluation on the external Messidor-2 dataset, which is a reassuring sign that our results generalize across datasets. 
The best-performing method across all training set fraction is IFI-CL, i.e.\ finetuning a model that was trained in a self-supervised fashion on ImageNet data, closely followed by IFI-FS, corresponding to the standard training methodology in medical imaging, where a model pretrained on ImageNet is finetuned on the target dataset. The results for 
the IWNE-CL model, i.e. self-supervised pretraining in target (DR) domain are weaker than the former two results.
This trend is again followed
at lower training set fractions where the model 
is trained with reduced fractions of the labeled dataset. 
While IWNE models deteriorate in performance,
IFI models show only a marginal drop
as shown in Figure~\ref{fig:auc}.

The results clearly advocate the use of IFI models as opposed to not using external data, which is in line with most part of the medical imaging literature but at first sight contradicts \cite{raghu2019transfusion}, who found no improvements from IFI as compared to direct training on a considerably larger closed source DR dataset. 
The inferior results of IWNE-CL compared to IFI-CL can potentially be attributed to two factors: the size of Eyepacs-1 as pretraining is with around 30k samples, very small compared to large natural image datasets, such as ImageNet with 1.2M images, where self-supervised contrastive methods were demonstrated to work really well. In addition, for IWNE-CL we used the same set of transformations proposed for ImageNet in \cite{chen2020simple}, which certainly represents a suboptimal choice for the DR images that differ qualitatively from natural images and the pretraining algorithm is rather sensitive to this choice.


\subsection{Statistics of Eigenvalues}
\label{sec:eignevalues}
\subsubsection{Condition number}
To better understand what
makes the IFI models 
achieve higher performance, 
we study the activations 
of the hidden layers. 
In particular, 
we compute the eigenvalues of 
the activations of
each layer in the four 
models we considered. 
Using the eigenvalues, we plot 
the condition number \cite{Goodfellow-et-al-2016}
as shown in Figure~\ref{fig:condition_number}.
To prevent the condition number 
from having very large values
due to division by the 
minimum of the eigenvalues,
we define the
condition number as follows: 
\begin{equation}
\kappa (A) = \frac{| \lambda_{p_{99.9}} (A)|}{| \lambda_{p_{90}} (A)|}  
\end{equation}
where $A$ are the activations 
of a hidden layer, 
$\kappa (A)$ is the condition number and 
$\lambda_{p_{i}} (A)$ is the eigenvalue 
corresponding to the ${i}^{th}$ percentile
of the eigenvalues.
While the top row in Figure~\ref{fig:condition_number}
shows the condition numbers
of the IWNE models, 
the bottom row shows the 
condition number of the 
IFI models.
The x-axis in both the figures 
corresponds to the layers of ResNet50. 

We find in Figure~\ref{fig:condition_number}
that the condition number for 
IFI models is much lower
than that of IWNE
implying significantly more
diverse features learned.
Also, in both versions of initializations, 
we find that the condition number for 
self-supervised learning is 
lower than that of supervised 
learning in the initial layers. 
This indicates that self-supervised 
learning extracts more diverse features 
than its supervised counterparts. 
We also find in Figure~\ref{fig:condition_number}
that for all the different models, 
the condition number is flattened out 
and becomes indistinguishable
for the latter layers. 
The initial layers form the crux of 
the learning process 
extracting effective and diverse feature representations
while the latter layers 
learn to aggregate these features.

\begin{figure*}[!t]
	\centering
	\resizebox{\textwidth}{!}{
    \begin{tabular}{ccccc}
		\toprule
		\hfil \textbf{Input} &
		\hfil  Microaneurysms & 
		\hfil Hemorrhages & 
		\hfil Hard Exudates & 
		\hfil Total\\
		\hline
 		\hfil{\includegraphics[width=0.2\textwidth, height=2.5in,,keepaspectratio]{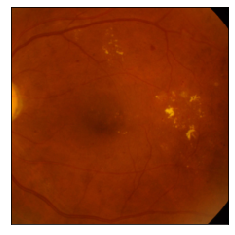}} &
 		\hfil{\includegraphics[width=0.2\textwidth, height=2.5in,,keepaspectratio]{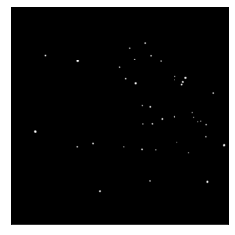}} &
		\hfil{\includegraphics[width=0.2\textwidth, height=2.5in,,keepaspectratio]{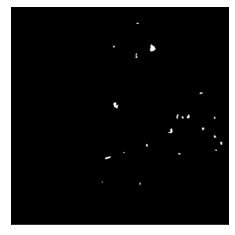}} &       \hfil{\includegraphics[width=0.2\textwidth, height=2.5in,,keepaspectratio]{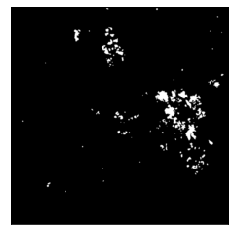}} &     
		\hfil{\setlength{\fboxrule}{3pt}
        \setlength{\fboxsep}{0pt}
        \fcolorbox{red}{red}{\includegraphics[width=0.2\textwidth, height=2.5in,,keepaspectratio]{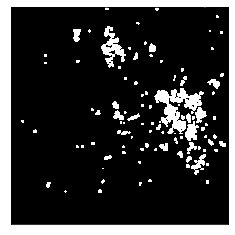}}}  \\
		\midrule
		\hfil \textbf{Methods} & \hfil IWNE-FS &  \hfil IWNE-CL &  \hfil IFI-FS &  \hfil IFI-CL\\
		\hline
		\noindent\parbox[m]{\hsize} {} &
		\hfil\raisebox{-.1\height}{\includegraphics[width=0.2\textwidth, height=2.5in,,keepaspectratio]{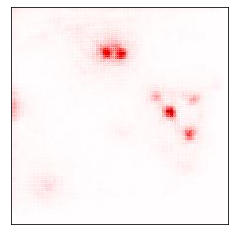}}  & 
		\hfil\raisebox{-.1\height}{\includegraphics[width=0.2\textwidth, height=2.5in,,keepaspectratio]{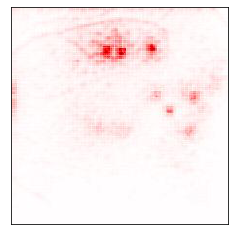}}  &
		\hfil\raisebox{-.1\height}{\includegraphics[width=0.2\textwidth, height=2.5in,,keepaspectratio]{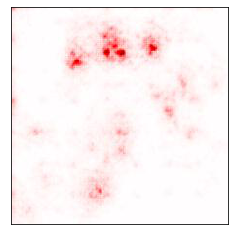}} & 
		\hfil\raisebox{-.1\height}{\includegraphics[width=0.2\textwidth, height=2.5in,,keepaspectratio]{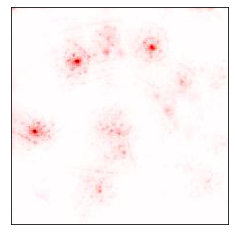}} 
		\\
		\bottomrule
	\end{tabular}
	}
	\caption{Top left image in the figure shows the input followed by the segmentation maps from the IDRiD dataset. 
	Top right image is the total which we compute by combining the segmentation maps of different lesions.
	Bottom row shows the explanation heatmaps for the given input.
	Each explanation heatmap is correlated with the total image marked in red to evaluate the effectiveness of the 
	model towards making the prediction for the disease.
	We find that IWNE-FS overfits on the hard exudates and also fails to pick up on cues related to microaneurysms. 
	We also find that explanation heatmaps
	of IFI models show reduced signs of overfitting to a single lesion when compared to IWNE. }
	\label{fig:idrid_1}
\end{figure*}

\begin{table*}
\centering
\resizebox{\textwidth}{!}
{
\begin{tabular}{@{}llllllllllll@{}}
\toprule
    \multirow{3}{*}{Lesions} &
    \multirow{3}{*}{Method} &
    \multirow{3}{*}{Pooling} &  
    \multicolumn{4}{c}{RMA} & 
    \multicolumn{4}{c}{RRA} \\

\cline{4-11}
    & & & 
    \multicolumn{2}{c}{Random} & 
    \multicolumn{2}{c}{LRP-$\alpha_{1}\beta_{0}$} & %
    \multicolumn{2}{c}{Random} & 
    \multicolumn{2}{c}{LRP-$\alpha_{1}\beta_{0}$} \\ %
\cline{4-11}
 & & & Mean & Median & Mean & Median & Mean & Median & Mean & Median  \\
\midrule
\multirow{8}{*}{Microaneurysms} & \multirow{2}{*}{IWNE-FS} 
 & sum\_pos    & 0.0073 & 0.0064 & 0.0076 & 0.0072 & 0.9885 & 0.9895 & 0.3447 & 0.4146\\
& & l2\_norm\_sq  & 0.0074 & 0.0067 & 0.0041 & 0.0025 & 0.9894 & 0.9900 & 0.4888 & 0.5047\\
 & \multirow{2}{*}{IWNE-CL } 
& sum\_pos    & 0.0074 & 0.0064 & \textbf{0.0093} & \textbf{0.0077} & 0.9879 & 0.9901 & \textbf{0.4370} & \textbf{0.5224} \\
& & l2\_norm\_sq  & 0.0073 & 0.0064 & \textbf{0.0075} & \textbf{0.0042} & 0.9882 & 0.9912 &  \textbf{0.5777} & \textbf{0.5736}\\
\cline{2-11}
 & \multirow{2}{*}{IFI-FS} 
& sum\_pos    & 0.0073 & 0.0061 & 0.0172 & 0.0143 & 0.9913 & 0.9922 & 0.5097 & 0.5551\\
& & l2\_norm\_sq  & 0.0074 & 0.0068 & 0.0374 & 0.0218 & 0.9891 & 0.9900 & 0.5705 & 0.5713\\
 & \multirow{2}{*}{IFI-CL} 
    & sum\_pos    & 0.0073 & 0.0061 & \textbf{0.0198} & \textbf{0.0186} & 0.9896 & 0.9897 &  \textbf{0.5831} & \textbf{0.6251} \\
& & l2\_norm\_sq  & 0.0073 & 0.0067 & \textbf{0.0595} & \textbf{0.0381} & 0.9902 & 0.9917 & \textbf{0.6357} & \textbf{0.6366}\\
\hline
\multirow{8}{*}{Haemorrhages} & \multirow{2}{*}{IWNE-FS} 
& sum\_pos    & 0.0234 & 0.0130 & 0.0251 & 0.0165 & 0.9902 & 0.9911 & 0.3845 & 0.4547\\
& & l2\_norm\_sq  & 0.0233 & 0.0126 & 0.0139 & 0.0056 & 0.9880 & 0.9905 &  0.5414 & 0.5565 \\
 & \multirow{2}{*}{IWNE-CL} 
& sum\_pos    & 0.0232 & 0.0126 &  \textbf{0.0602} & \textbf{0.0458}  & 0.9889 & 0.9904 & \textbf{0.4971} & \textbf{0.6076}\\
& & l2\_norm\_sq  & 0.0234 & 0.0125 & \textbf{0.1063} & \textbf{0.0525} & 0.9881 & 0.9892 &  \textbf{0.6357} & \textbf{0.6371}\\
\cline{2-11}
 & \multirow{2}{*}{IFI-FS} 
& sum\_pos    & 0.0233 & 0.0126 & 0.0711 & 0.0578 & 0.9896 & 0.9895 &  0.5840 & 0.6127\\
& & l2\_norm\_sq  & 0.0233 & 0.0125 & 0.1438 & 0.1243 & 0.9891 & 0.9904 &  0.6571 & 0.6551\\
 & \multirow{2}{*}{IFI-CL} 
& sum\_pos    & 0.0234 & 0.0127 & \textbf{0.0765} & \textbf{0.0722 } & 0.9897 & 0.9911 & \textbf{0.6898} & \textbf{0.7194}\\
& & l2\_norm\_sq  & 0.0234 & 0.0125 & \textbf{0.1874} & \textbf{0.1808} & 0.9873 & 0.9911 & \textbf{0.7403} & \textbf{0.7405} \\

\hline
\multirow{8}{*}{Hard Exudates} & \multirow{2}{*}{IWNE-FS} 
& sum\_pos    & 0.0409 & 0.0195 &  \textbf{0.1954} & \textbf{0.1734} & 0.9897 & 0.9906 &  0.5086 & \textbf{0.6959}\\
& & l2\_norm\_sq  & 0.0409 & 0.0190 & \textbf{0.4201} & \textbf{0.4921} & 0.9898 & 0.9903 &  \textbf{0.7114} & \textbf{0.7435}\\
 & \multirow{2}{*}{IWNE-CL} 
& sum\_pos    & 0.0409 & 0.0200 & 0.1206 & 0.1018 & 0.9889 & 0.9898 & \textbf{0.5136} & 0.5887\\
& & l2\_norm\_sq  & 0.0409 & 0.0191 & 0.2338 & 0.1652 & 0.9892 & 0.9898 & 0.6656 & 0.7038 \\
\cline{2-11}
 & \multirow{2}{*}{IFI-FS} 
& sum\_pos    & 0.0408 & 0.0195 & \textbf{0.1103} & \textbf{0.0861} & 0.9895 & 0.9905 &  \textbf{0.5480} & \textbf{0.6258} \\
& & l2\_norm\_sq  & 0.0410 & 0.0194 & \textbf{0.2125} & \textbf{0.1659} & 0.9905 & 0.9915 & \textbf{0.6125} & \textbf{0.6561}\\
 & \multirow{2}{*}{IFI-CL} 
& sum\_pos    & 0.0409 & 0.0188 &  0.0725 & 0.0533  & 0.9890 & 0.9896 & 0.5425 & 0.5762\\
& & l2\_norm\_sq  & 0.0412 & 0.0193 & 0.1195 & 0.0858 & 0.9888 & 0.9903 & 0.6088 & 0.6260\\
\hline
\multirow{8}{*}{Total} & \multirow{2}{*}{IWNE-FS} 
& sum\_pos    & 0.0710 & 0.0558 &  \textbf{0.2266} & \textbf{0.2000} & 0.9899 & 0.9905 &  0.4459 & 0.5655 \\
& & l2\_norm\_sq  & 0.0711 & 0.0563 &  \textbf{0.4363} & \textbf{0.5104}  & 0.9893 & 0.9898 &  0.6151 & \textbf{0.6503}\\
 & \multirow{2}{*}{IWNE-CL} 
& sum\_pos    & 0.0710 & 0.0565 & 0.1887 & 0.1619 & 0.9884 & 0.9891 &  \textbf{0.5015} & \textbf{0.6083}\\
& & l2\_norm\_sq  & 0.0711 & 0.0565 & 0.3457 & 0.3330 & 0.9884 & 0.9890 & \textbf{0.6459} & 0.6334\\
\cline{2-11}
 & \multirow{2}{*}{IFI-FS} 
& sum\_pos    & 0.0709 & 0.0561 & \textbf{0.1969} & \textbf{0.1886} & 0.9893 & 0.9897 & 0.5479 & 0.5941 \\
& & l2\_norm\_sq  & 0.0711 & 0.0557 &  \textbf{0.3905} & \textbf{0.3964} & 0.9893 & 0.9896 & 0.6150 & 0.6245\\
 & \multirow{2}{*}{IFI-CL} 
& sum\_pos    & 0.0711 & 0.0576 & 0.1671 & 0.1724 & 0.9893 & 0.9896 & \textbf{0.5847} & \textbf{0.6144} \\
& & l2\_norm\_sq  & 0.0713 & 0.0569 & 0.3625 & 0.3650 & 0.9895 & 0.9897 & \textbf{0.6428} & \textbf{0.6463} \\
\bottomrule
\end{tabular}
}
\caption{Relevance mass accuracy (RMA) 
and relevance rank accuracy (RRA)
on 
the LRP-$\alpha_{1}\beta_{0}$ explanation heatmaps of images
of the IDRiD dataset. 
The results show that 
while supervised models overfit on the hard exudates,
the self-supervised models 
look at diverse set of input features (lesions). 
On the other hand, we also find that 
IFI models show higher accuracies when compared
to IWNE models. 
}
\label{tab:accuracy_all}
\end{table*}

\subsubsection{Spread of Eigenvalues}
To investigate 
the distinctive aspects of the
initial layers, 
we plot the eigenvalues of the first layer
for all four models in Figure~\ref{fig:eigenvalues_conv1}. 
The eigenvalues are made symmetrical 
around $0$ and plotted in the form of density
to make for better visualization. 
The bottom row in Figure~\ref{fig:eigenvalues_conv1} 
also zooms in on the tails. 
We find that the 
IWNE models obtain high and peaked eigenvalues.
On the other hand,
the IFI models, 
show lower peak values.
Similar to the findings 
in the experiments on the
condition number, 
self-supervised learning
in contrast to supervised learning 
shows a slightly lower peak value. 
Additionally, in both versions 
of the initialization, self-supervised learning
models show more 
heavy tailedness. 

The results indicate that 
IWNE models
learn kernels 
in the first convolutional layer
that are activated for some very specific patterns.
On the contrary,
IFI models 
learn kernels 
that activate for a broader range of 
input features. 
The superior performance of IFI models 
can be attributed to this effect while
this may lead to several other
benefits including
increase in generalization
and robustness.

\subsubsection{Distribution Fitting}
In this section, 
we fit the eigenvalues 
of the first convolutional layer
to the parameters of several
distributions and report the 
distribution that fits best
\cite{erdogant2019distfit}.
Among a wide range parameterized distributions, 
we find in Table~\ref{tab:distfit} that 
all the four models fit best to 
the \emph{Pareto} distribution,
though the parameters vary. 
Pareto distribution with 
the shape parameter $\alpha=1.16$ corresponds to the $80-20$ rule,
implying that $80\%$ of the results
come from $20\%$ of the causes \cite{pareto1964cours}. 
IWNE models 
show $\alpha$ values higher than $1.16$. 
This indicates that 
the overall result comes from 
less than $20\%$ of the
activations.  
In other words, 
the kernels learned by the 
IWNE models 
extract small number of, 
yet highly curated set of features
from the input. 
In contrast, we find that 
IFI brings down the value
of $\alpha$ for the Pareto distribution
implying a wider range of feature representations
learned by the first convolutional layer. 
Additionally, 
in both versions of initializations,
CL shows reduced value of $\alpha$ 
when compared to FS indicating
that the kernels learned by CL methods 
fire on a further broader range of input.

Our studies show that 
pretraining and self-supervised learning 
is beneficial for the downstream 
medical imaging 
task to be able learn kernels 
that fire broadly and in turn
extract more diverse and effective 
features from the input.

\subsection{Robustness to Distortions:}
\label{ssec:robustnesstodistortions}

The heavy-tailed activation statistics in combination with ReLU-thresholding in Section \ref{sec:eignevalues} showed that a larger number of neurons are capable of detecting structures in the input when the input data is varied according to sampling from the dataset. One can expect that this also may translate to an increased detection capability when input samples are varied by data augmentation parameters towards zones of lower data density. We have performed this experiment for the IWNE and IFI models 
by distorting the input with a set of predefined distortions as shown in \cite{hendrycks2019robustness}.

One can see from Figure~\ref{fig:robustness} that for the majority of distortion cases, the score for the self-supervised model is higher, indicating a higher robustness to the respective distortions. There is a marked difference between
IWNE and IFI models. 
In the former case CL always provides an increase in robustness
in comparison to FS.
Using IFI in the latter case is known to provide good generalization for finetuning with respect to a wide range of target datasets. This improved generalization levels the difference between FS and CL.
However IFI-CL still improves robustness for different noise types, pixelation and lower levels of saturation changes. Note the conspicuous outlier in IFI for JPEG compression. 

\subsection{Quantitative Analysis of Learned Cues} 
\label{sec:idrid}
Explainability for DNN reveals what the model looks at on the image to make the prediction \cite{BachPLOS15, SamTNNLS17, MonDSP18, SamITU18b, sundararajan2017axiomatic, shrikumar2017learning,samek2021toward, SamXAI19, hagele2020resolving, holzinger2019causability, holzinger2020artificial, holzinger2021towards}. 
Using ground-truth segmentation masks, 
explanations have been evaluated to show quantitatively 
if what the model is looking at,
is relevant for making the decision \cite{OsmArXiv20}. 
In the case of DR, 
a reasonable expectation is that
the trained model looks at 
lesions in the retina that is 
indicative of the disease in order 
to make its decision.
In order to evaluate the explanation 
heatmaps,
we use the dataset of IDRiD \cite{porwal2020idrid}
containing detailed pixel-wise annotation 
of the different lesions that contribute to the disease. 
The dataset consists of $80$ images\footnote{The IDRiD dataset also contains segmentation maps for soft exudates for a smaller subset of images which we excluded from our quantitative evaluation.} with segmentation masks for microaneurysms, 
haemorrhages and hard exudates. To obtain explanation heatmaps, we utilize Layer-wise Relevance Propagation (LRP) with $\alpha_{1}\beta_{0}$ rule \cite{MonDSP18,samek2021toward}.

Figure~\ref{fig:idrid_1} shows the input followed by the 
segmentation maps for different lesions in the top row. 
The final image in the top row combines the different lesions to
form the total. 
The bottom row shows the explanation heatmaps 
by using the different training methods. 
By comparing each result to the total marked in red
in Figure~\ref{fig:idrid_1}, 
we can evaluate the effectiveness of the model
in looking at the lesion to make the prediction. 
We find that explanation heatmaps from IWNE
overfit on the hard exudates 
and show minimal correlation with the other lesions. 
On the other hand, 
explanation heatmaps from IFI models 
are significantly more outspread 
correlating better with different lesions. 

The correlation of
explanation heatmaps to the 
ground-truth segmentation maps also
helps us make a quantitative
evaluation of how accurately 
the models relies on the disease to make its prediction.
We follow the evaluating strategies adopted in \cite{OsmArXiv20} including relevance mass accuracy and relevance rank accuracy. Given input $\boldsymbol{x}$, 
relevances $R_{i}$ determining the importance of the input features $x_{i}$ and $S \subseteq [0,1]$ the ground truth segmentation mask, relevance mass accuracy is defined as: 
\begin{equation}
\label{eq:rma}
	\text{RMA} = \frac{\sum_{i\in S}R_{i}}{\sum_{i}R_{i}}
\end{equation}
where the numerator corresponds to
the sum of relevances where the ground truth segmentation maps exists and the denominator is the sum of all relevances. 
The relevance rank accuracy is defined as:
\begin{equation}
\label{eq:rra}
	\text{RRA} = \frac{|{R_{p_{i}}} \cap S|}{|S|}
\end{equation}
where $R_{p_{i}}$ is the relevances in the top $i^{th}$ percentile. 
While RMA corresponds to the precision, 
RRA corresponds to the recall. 
For pooling the relevances across the channels, 
we utilize the two pooling strategies 
followed by 
\cite{OsmArXiv20}, although the findings here
are agnostic to the pooling strategy utilized: 
\begin{itemize}
	\item $\text{sum\_pos:} \;  R_{pool}     = max(0,\sum_{i=1}^{C}R_i) $   
	\item     $\text{l2\_norm\_sq:} \;   R_{pool} = \sum_{i=1}^{C}{R_i}^2 $
	\end{itemize}
where $C$ is the number of channels. 

Table~\ref{tab:accuracy_all} shows the results for RMA and 
RRA for the explanation heatmaps correlated on the ground-truth segmentation maps from the IDRiD challenge. 
We report the accuracies for each lesion---microaneurysms, 
haemorrhages and hard exudates and a total, where we combine the above mentioned lesions.
The heatmaps for each of the methods 
are computed by backpropagating 
from the output neuron
corresponding to severe DR.
The heatmaps are 
evaluated using the two pooling strategies
mentioned above for each lesion. 
As a control, we also report 
the results by replacing
explanation heatmaps with
random variables from Gaussian distribution. 
Any method which shows similar
results to the control indicates that 
the heatmaps are just random, 
i.e.\ the model looks at random set
of input features to make its prediction.
In each category (lesion), the best result among 
the different training strategies 
are marked in bold for each pooling method.

We find in Table~\ref{tab:accuracy_all} that
in the case of microaneurysms, 
random explanations achieve a mean 
accuracy of $0.0073$ for RMA. 
Here, the model IWNE-FS achieves results 
that is very close to the results for 
the random explanations. On the other hand, 
all the other models report accuracies that are higher than the corresponding control value. 
This indicates that IWNE-FS 
may be ignoring 
microaneurysms for making its decision.
The RMA results in
Table~\ref{tab:accuracy_all} show that
for the IWNE models,
CL achieves better results.
IFI models, in general report
higher accuracies than that of IWNE models. 
Similar to IWNE, we find for
IFI models that
CL reports better RMA than FS
using both the pooling strategies. 
This is confirmed again with 
results of RRA in the same table,
where models with
CL achieves the 
best results.
Microaneurysms are the smallest lesions 
and it is vital for a method to 
base its decision on them for detecting
progressive cases of DR. 
Our results indicate that
IFI models and CL in particular
are better equipped at 
including microaneurysms
to make their predictions.

Haemorrhages are lesions
that are slightly larger than microaneurysms. 
We find in Table~\ref{tab:accuracy_all} that 
here again IWNE-FS reports similar accuracies 
to that of the control indicating that
this model may be ignoring the haemorrhages as well. 
Among IWNE models,
CL clearly achieves higher RMA as well as higher RRA. 
This is again the case on the 
IFI models where
CL achieves higher RMA and RRA indicating that 
the explanations using this model
are better correlated with 
the ground-truth than 
their supervised counterpart FS.

In contrast to the smaller lesions, 
the hard exudates are large yellowish white deposits 
with sharp gradients. 
Here for RMA, 
the supervised models achieve better results
than the self-supervised models
as shown in Table~\ref{tab:accuracy_all}.
The results on RRA for hard exudates show 
that on majority of the cases, 
for both IWNE and IFI models,
the supervised models show higher accuracies than
the self-supervised models. 

The \emph{total} which measures the sum of the all the different 
lesions, we find here again that 
the supervised models
achieve better results with RMA as shown in 
~\ref{tab:accuracy_all}.
With RRA, 
the IWNE
models do not clearly outperform
each other in the case of total. 
However, for IFI, 
the self-supervised model clearly outperforms the 
supervised model for the total 
of all the lesions. 



The results of RMA and RRA in 
Table~\ref{tab:accuracy_all} 
reveal that the supervised models 
overfit on the hard exudates 
in both versions of initializations.
IWNE-FS in particular 
fails to base its decision on   
microaneurysms and haemorrhages
that may be highly relevant 
for the prediction of the disease. 
The results on the total are skewed by the 
results on the hard exudates.
In alignment with our observations in 
Section~\ref{sec:eignevalues}, 
we find that
the IFI models
look at diverse set of 
input features (lesions)
and report consistently 
higher accuracies than their 
IWNE counterparts. 
Among IFI, 
the results of CL 
correlates better with the 
explanation heatmaps for a variety of lesions
indicating that they look at more 
diverse set of input features than any other method.

\section{Summary and conclusions} 
Deep learning-based methods for the diagnosis of diabetic retinopathy have shown remarkable performance. 
In our paper, 
we study the important question of 
the robustness of different training strategies ---
namely initialization from ImageNet pretraining and self-supervised learning. 
Our findings are three-fold: 
Firstly, we show the performance gains obtained by self-supervised learning in diabetic retinopathy.
Secondly, we demonstrate the advantage of self-supervised learning along with initialization from ImageNet pretraining for diabetic retinopathy by analyzing 
the statistics of the eigenvalues of the 
feature representations learned.
We also show improvements in 
robustness to distortions for 
self-supervised learning 
in comparison to purely supervised training.
Finally, we use interpretability methods to gain quantitative insights into the patterns exploited by models trained using the different training schemes. In particular, we find that initialization from ImageNet pretraining significantly 
reduces overfitting to large lesions along with improvements in taking into account minute lesions which are indicative of the progression of the disease. 

With our study, we try to convey that a more holistic view on the benefits of pretraining and self-supervision in medical imaging along the lines of the present study is important. To summarize, in absence of large unlabeled domain-specific data that would allow for self-supervised pretraining, we see numerous benefits in favor of using self-supervised pretrained models on Imagenet as starting point for finetuning on domain-specific data, which we put as a general recommendation.

\section*{Acknowledgments}
Models were trained in Pytorch \cite{PytorchNIPS2019} building on the code kindly provided by \cite{he2020momentum}. 
This work was supported in part by the German Ministry for Education and Research (BMBF)
under grants 01IS14013A-E, 01GQ1115, 01GQ0850, 01IS18056A, 01IS18025A and 01IS18037A. 
It is also supported in part by the
Information \& Communications Technology Planning \& Evaluation (IITP) grant funded by the Korea government (No. 2017-0-001779), 
as well as by Math+, EXC 2046/1, Project ID 390685689 through the German Research Foundation (DFG).
Correspondence to WS, KRM. 

\bibliographystyle{IEEEtran}
\bibliography{bibfile}

\end{document}